\documentclass[11pt]{article}

\usepackage[margin=1in]{geometry}
\usepackage[T1]{fontenc}
\usepackage[utf8]{inputenc}
\usepackage{lmodern}
\usepackage{microtype}
\usepackage{booktabs}
\usepackage{array}
\usepackage{longtable}
\usepackage{tabularx}
\usepackage{enumitem}
\usepackage{amsmath}
\usepackage{amssymb}
\usepackage{xcolor}
\usepackage{url}
\usepackage{hyperref}
\usepackage[numbers,sort&compress]{natbib}

\hypersetup{
  colorlinks=true,
  linkcolor=blue,
  citecolor=blue,
  urlcolor=blue
}

\setlist[itemize]{leftmargin=1.3em}
\title{RACL: Reasoning-Agent Control Layers for Continuous Metaheuristic Learning}

\author{
  Antón Asla Manzárraga\\
  Independent Researcher\\
  \texttt{amanzarraga@gmail.com}\\
  ORCID: \href{https://orcid.org/0009-0008-0242-7186}{0009-0008-0242-7186}
}

\date{}

\begin{document}

\maketitle

\begin{abstract}
Metaheuristic optimization is widely used in operational decision systems, but effective long-term use often requires expertise that many companies do not have internally. A company may have access to a configured optimizer while lacking the optimization expertise required to adapt its internal search behavior as operational experience accumulates.

This paper introduces \textbf{RACL}, a \textbf{Reasoning-Agent Control Layer} for metaheuristics. RACL places a reasoning agent above an existing optimizer. The agent does not replace the optimizer and does not modify business constraints. Instead, it learns how to control the optimizer's internal search behavior by observing operational memory, reasoning over past behavior, formulating bounded hypotheses, testing interventions, evaluating outcomes, applying guardrails, consolidating useful policies and explaining its decisions.

The experiment uses vehicle routing as a testbed, but the contribution is not a new routing solver, a particular ALNS configuration or a specific set of routing rules. The contribution is the RACL method: a way for a reasoning agent to discover, validate, consolidate and explain algorithmic control rules for a metaheuristic.

In the current experimental setting, RACL improves or ties the Operational Memory Policy in 21 of 21 feasible cases and improves or ties a non-reasoning Stagnation-Triggered Policy in 18 of 21 feasible cases, with an average RACL vs STP cost delta of -0.641\%. In the Sevilla-9/10 runtime sample, RACL improves average cost by -8.337\% versus Fixed and -1.605\% versus STP without showing material computational overhead.

During the proof-of-concept, Codex was used as an in-the-loop reasoning agent observing executions, interpreting logs and proposing live bounded interventions. The policy proxy was later used only to make quantitative evaluation reproducible. The results support the central claim: reasoning agents can act as continuous control-learning layers for metaheuristics, generating useful and explainable algorithmic control behavior without taking ownership of business constraints.
\end{abstract}

\section{Introduction}

Many operational optimization systems are used repeatedly in similar contexts. Logistics is a clear example: companies may solve routing problems every day in the same geographic area, with changing demand but recurring operational structure. Metaheuristics are attractive in these settings because they can produce good solutions under complex constraints where exact optimization may be impractical.

However, a gap remains between having an optimizer and improving it over time. In many deployments, the optimization engine is configured once and then used repeatedly. The client defines business constraints such as fleet availability, delivery priorities, time limits and service rules, but may not have optimization experts who can continuously adjust the internal behavior of the metaheuristic when performance stagnates or when recurring patterns appear in the operation.

This paper addresses that gap by proposing RACL, a Reasoning-Agent Control Layer. RACL places a reasoning agent above an existing metaheuristic. The agent observes operational memory, reasons about search behavior, proposes bounded algorithmic interventions, evaluates their effects, applies guardrails and consolidates useful policies. The agent does not alter business constraints. Its role is narrower and more technical: it controls how the metaheuristic searches.

The motivating scenario is a company that repeatedly solves similar optimization problems without having an internal optimization specialist. In that setting, every execution becomes evidence. RACL turns this accumulated operational memory into a resource for continuous algorithmic improvement.

The paper proposes a memory-guided reasoning cycle:

\begin{center}
\small
\begin{tabular}{c}
\texttt{operational memory $\rightarrow$ reasoning $\rightarrow$ hypothesis $\rightarrow$ bounded experiment} \\
\texttt{$\rightarrow$ guardrail $\rightarrow$ policy consolidation $\rightarrow$ business explanation $\rightarrow$ memory update}
\end{tabular}
\end{center}

The routing experiment in this paper is a testbed for this idea. The specific ALNS-style engine, city and discovered rules are not the core contribution. They are evidence that the RACL process can produce meaningful control behavior.

The central research question is:

\begin{quote}
Can a reasoning agent learn useful metaheuristic control behavior from operational memory and bounded experimentation?
\end{quote}

The results support a positive answer. RACL produces rules that are useful in the testbed, but the rules themselves are not the main result. The main result is the method by which the rules are obtained, tested, guarded, consolidated and explained.

\section{Related Work}

This work sits at the intersection of hyper-heuristics, adaptive operator selection, case-based reasoning, learning-based optimization and LLM-assisted heuristic generation.

Hyper-heuristics are the closest methodological family. Early work framed hyper-heuristics as high-level methods that select or generate heuristics rather than directly solving only in the solution space \citep{Burke2003HyperHeuristicsEmerging}. Later surveys consolidated this view and positioned hyper-heuristics as a way to raise the level of generality in heuristic search \citep{Burke2013HyperHeuristicsSurvey}. Selection hyper-heuristics are particularly relevant because they choose among existing low-level heuristics during the search \citep{Drake2020SelectionHyperHeuristics}. RACL shares this high-level control view, but differs in that the controller is explicitly framed as a reasoning agent that formulates, tests, consolidates and explains hypotheses using operational memory.

Adaptive Large Neighborhood Search (ALNS) provides the experimental base used in this paper, but RACL is not specific to ALNS. Ropke and Pisinger applied ALNS to pickup-and-delivery problems with time windows \citep{Ropke2006ALNSPickupDelivery}, and later work summarized large-neighborhood search and ALNS principles as a general metaheuristic framework \citep{Pisinger2010LargeNeighborhoodSearch}. Recent work has explored more advanced online control of ALNS, including deep reinforcement learning for operator and parameter control \citep{Reijnen2022DRALNS} and graph reinforcement learning for operator selection \citep{Johnn2023GraphRLALNS}. These approaches are close because they observe the search process and adapt algorithmic behavior. RACL differs by treating control as a reasoning process over operational memory: the agent does not merely select an operator; it explains, tests, guards and consolidates control knowledge.

Case-based reasoning informs the memory component. Classical CBR retrieves, reuses, revises and retains prior cases \citep{Aamodt1994CBR,Kolodner1993CBR}. RACL follows this intuition, but stores search-control experience rather than complete routing solutions. Each case may include the observed search state, selected action, justification, outcome and feasibility status. Memory is therefore treated as evidence for control decisions, not as a repository of route plans to replay.

Learning-based optimization has grown substantially. Surveys cover both general machine-learning approaches for combinatorial optimization \citep{Bengio2021MLCO} and VRP-specific learning methods \citep{Li2021LearningVRPOverview}. Neural combinatorial optimization with reinforcement learning is an important early example of learning policies for routing-like problems \citep{Bello2016NeuralCombinatorialOptimization}. RACL is closer to hybrid or step-by-step learning-based optimization than to end-to-end neural routing: the agent does not output routes directly, but controls a traditional optimizer that remains responsible for route construction and feasibility.

Recent LLM-based work has shown that language models can participate in heuristic generation, program search and automatic algorithm design. FunSearch combines LLM-generated program variants with an evaluator and evolutionary search \citep{RomeraParedes2024FunSearch}. ReEvo frames LLMs as language hyper-heuristics with reflective evolution \citep{Ye2024ReEvo}. LLaMEA uses an LLM-driven evolutionary loop to generate metaheuristics \citep{VanStein2024LLaMEA}, while Evolution of Heuristics uses LLMs to evolve natural-language heuristic ideas and executable code \citep{Liu2024EvolutionOfHeuristics}. These systems use language-based reasoning to improve optimization behavior. RACL takes a different position: the agent does not generate a new solver or new heuristic code; it governs an existing metaheuristic through bounded, auditable control actions.

The research gap addressed here is operational. The question is not only whether a model can design a heuristic, but whether a reasoning agent can act as a practical control-learning layer over an existing optimization engine, using historical local evidence, preserving customer-owned business constraints and producing explanations that non-technical operational users can understand.

\section{RACL Method}

\subsection{Overview}

RACL treats the reasoning agent as a control-learning layer for a metaheuristic optimizer.

The optimizer remains responsible for producing feasible solutions. The agent controls the search process by selecting bounded algorithmic actions. It cannot modify delivery requirements, fleet constraints, priority deadlines or any other business rule.

The RACL cycle is:

\begin{center}
\texttt{observe $\rightarrow$ retrieve $\rightarrow$ reason $\rightarrow$ hypothesize $\rightarrow$ intervene $\rightarrow$ evaluate $\rightarrow$ guard $\rightarrow$ consolidate $\rightarrow$ explain $\rightarrow$ update memory}
\end{center}

\subsection{Full Architecture and Current Realization}

In the full RACL architecture, the reasoning agent is an online layer. It observes each execution block, retrieves operational memory, applies consolidated policies when appropriate, formulates new hypotheses when the current situation is not well covered, tests bounded interventions, evaluates their effects, applies guardrails and updates memory over time.

The current experimental realization does not require a production LLM service. During the proof-of-concept, Codex was used as an in-the-loop reasoning agent observing executions, interpreting logs and proposing live bounded interventions. It reviewed states and logs, used accumulated memory, proposed hypotheses, selected bounded interventions, interpreted outcomes and introduced guardrails. The resulting RACL behavior was then encoded as a local policy proxy for reproducible evaluation against baselines.

This distinction is methodological. The paper validates the process by which the agent reaches useful control rules. The policy proxy is the evaluation artifact, not the scientific object.

\subsection{Runtime Reasoning and Memory Consolidation}

RACL is a runtime learning loop, not a static rule. In the target architecture, the optimizer executes a block, the agent receives the observable search state, retrieves relevant memory, decides whether to apply a consolidated policy or formulate a new bounded hypothesis, applies a controlled intervention, evaluates the outcome, introduces guardrails when risk appears and consolidates policies when evidence is consistent.

The loop can be summarized as a sequence of bounded runtime operations:

\begin{enumerate}[leftmargin=1.5em]
  \item observe the search state and retrieve operational memory;
  \item reason over matched evidence and current behavior;
  \item apply a matching policy or formulate a bounded hypothesis;
  \item intervene for a controlled block;
  \item evaluate cost, stagnation, degradation and feasibility;
  \item add a guardrail if risk appears;
  \item consolidate the policy if evidence is repeated;
  \item explain the decision and update memory for future executions.
\end{enumerate}

\begin{table}[t]
\centering
\caption{Runtime RACL cycle.}
\label{tab:racl-cycle}
\begin{tabular}{lll}
\toprule
Step & Runtime role & Evidence produced \\
\midrule
Observe & Read state after a block & Search-state summary \\
Retrieve & Consult memory & Matched prior evidence \\
Reason & Interpret state and memory & Diagnosis of opportunity or risk \\
Hypothesize & Propose bounded control idea & Decision rationale \\
Intervene & Apply algorithmic action & Action log \\
Evaluate & Check cost and feasibility & Outcome metrics \\
Guard & Limit risky behavior & Guardrail rule \\
Consolidate & Retain useful behavior & Updated policy memory \\
Explain & Translate decision to business language & Business-readable trace \\
\bottomrule
\end{tabular}
\end{table}

\subsection{Runtime RACL Decision Traces}

To make the reasoning layer observable, the experiment records intervention episodes with the state before the action, the action selected and the post-intervention outcome. Table~\ref{tab:racl-traces} shows representative traces from the Sevilla-9/10 validation episodes.

\begin{table}[t]
\centering
\caption{Representative runtime RACL decision traces.}
\label{tab:racl-traces}
\small
\begin{tabularx}{\linewidth}{lllp{0.26\linewidth}X}
\toprule
Dataset & Seed & Block & RACL decision & Outcome \\
\midrule
Sevilla-9 & 101 & 6 & \texttt{route\_probe} & 1667.862 to 1468.862 within two blocks (-11.931\%) \\
Sevilla-10 & 404 & 6--7 & \texttt{route\_diversify} + \texttt{intensify} & 1576.723 to 1417.664 within two blocks (-10.088\%) \\
Sevilla-10 & 101 & 10 & \texttt{route\_probe} & No additional improvement; retained as neutral evidence \\
\bottomrule
\end{tabularx}
\end{table}

These traces are important because they show RACL acting during the search, not only reporting an aggregate result at the end. A business-readable explanation of the second trace is: the planning process appeared stuck, so the agent allowed a stronger route reorganization; once a better region was found, it switched to a more focused search to stabilize the improvement, while feasibility checks continued to protect capacity, time and priority constraints.

\subsection{Observable State, Memory and Actions}

The agent observes a compact state summary after each block: best cost, current cost, recent improvement, acceptance rate, improving rate, iterations since last improvement, route count, unassigned routes and feasibility indicators.

Operational memory stores previous states, actions, reasons, outcomes and feasibility results. Memory is not treated as truth. It is evidence that the agent can retrieve, reinterpret and revise.

The current action space includes \texttt{baseline}, \texttt{route\_probe}, \texttt{return\_to\_baseline}, \texttt{route\_diversify} and \texttt{intensify\_after\_route\_diversify}. These actions are experimental examples of bounded algorithmic interventions. They are not universal rules and are not the contribution itself.

\subsection{Continuous Learning}

RACL is designed for repeated operational use. It is not expected to discover one rule and stop. Over time, the agent may reinforce rules that continue to work, narrow rules that only work under specific contexts, retire rules that stop working, formulate new hypotheses when memory suggests an opportunity, test new interventions safely and consolidate new policies when evidence supports them.

This continuous learning view is central to the paper. The value of RACL lies in the method for reaching and revising rules, not in the specific rules obtained in one experiment.

\section{Experimental Design}

\subsection{Purpose of the Experiment}

The experiment tests whether the RACL method can produce useful control behavior.

The routing engine, datasets and concrete control actions are an experimental vehicle. They are not the main contribution. The experiment is designed to answer whether an agent can use memory and reasoning to discover, test and consolidate algorithmic control rules that improve a baseline metaheuristic while preserving feasibility.

\subsection{Compared Strategies}

Four strategies are compared.

\textbf{Fixed Baseline} executes the routing metaheuristic without adaptive intervention.

\textbf{Operational Memory Policy (OMP)} represents the first policy derived from historical memory. It captures conservative memory-supported control.

\textbf{Stagnation-Triggered Policy (STP)} is a non-reasoning hyper-heuristic baseline. It reacts to stagnation by applying a bounded route-level probe and then returns to baseline. It does not use memory retrieval, hypothesis generation or learned guardrails.

\textbf{RACL policy proxy} represents the behavior consolidated by the reasoning-agent process. It is the evaluated artifact distilled from Codex-in-the-loop reasoning, bounded experimentation and guardrail consolidation.

\subsection{Metrics}

The main metric is final route cost. Feasibility is checked before treating any result as valid evidence.

The analysis also reports win/tie/loss counts, mean and median cost deltas, runtime, intervention counts, paired statistical tests, traceability of agent decisions and business-readable explanations.

\section{Results}

The results should be interpreted as evidence for the RACL method, not as claims about universal superiority of a specific rule.

\subsection{RACL vs STP}

STP is a useful baseline because it reacts to stagnation without memory or reasoning. Across 21 feasible cases, RACL is better than STP in 11 cases, tied in 7 cases and worse in 3 cases. RACL therefore improves or ties STP in 18 of 21 cases, with a mean cost delta of -0.641\%.

This result shows that the agentic cycle adds value beyond a simple non-reasoning trigger in aggregate.

\subsection{RACL vs OMP}

Across 21 feasible cases, RACL improves OMP in 18 cases and ties in 3, with no degradation. The mean RACL delta versus OMP is -4.913\%.

This supports the central methodological claim: reasoning over memory can produce richer useful control behavior than a first memory-derived rule.

\subsection{RACL vs Fixed Baseline}

In the Sevilla-9/10 paired sample where Fixed is available, RACL improves Fixed in 8 of 8 feasible cases, with an average cost improvement of -8.337\%.

This result supports the practical motivation: an agentic control layer can improve a configured optimizer without changing business constraints.

\subsection{Runtime}

Runtime was measured on Sevilla-9 and Sevilla-10 over four seeds. Table~\ref{tab:runtime} summarizes the measured sample.

\begin{table}[t]
\centering
\caption{Runtime and average improvement on the Sevilla-9/10 measured sample.}
\label{tab:runtime}
\begin{tabular}{lrrrr}
\toprule
Strategy & Cases & Avg. runtime (s) & Runtime ratio vs Fixed & Avg. improvement vs Fixed \\
\midrule
Fixed & 8 & 159.857 & 1.000 & 0.000\% \\
OMP & 8 & 156.828 & 0.982 & 0.000\% \\
STP & 8 & 137.043 & 0.865 & -6.752\% \\
RACL & 8 & 140.769 & 0.886 & -8.337\% \\
\bottomrule
\end{tabular}
\end{table}

RACL improves solution quality without showing material computational overhead in the measured cases.

\section{Statistical Analysis}

Paired tests are used because strategies are compared on the same dataset and seed. Table~\ref{tab:stats} summarizes the paired comparisons.

\begin{table}[t]
\centering
\caption{Paired comparison summary.}
\label{tab:stats}
\begin{tabular}{lrrrrrr}
\toprule
Comparison & Cases & Wins & Ties & Losses & Mean delta & Directional $p$ \\
\midrule
RACL vs STP & 21 & 11 & 7 & 3 & -0.641\% & 0.037 \\
RACL vs OMP & 21 & 18 & 3 & 0 & -4.913\% & 0.000098 \\
RACL vs Fixed & 8 & 8 & 0 & 0 & -8.337\% & 0.004 \\
\bottomrule
\end{tabular}
\end{table}

The strongest result is RACL vs OMP: the reasoning layer improves the first memory-derived policy. RACL vs Fixed is also strong in the paired sample. RACL vs STP should be read as favorable aggregate evidence rather than a claim of dominance, which is important because STP already captures part of the benefit of reacting to stagnation.

\section{Business-Facing Explainability}

RACL is designed for settings where non-technical users need to understand why the optimizer changed its behavior.

An example explanation is:

\begin{quote}
The optimizer had spent several blocks without finding better route plans. The agent first tried a moderate route reorganization. Because stagnation persisted and there was still enough search horizon to recover if the test failed, it tried a stronger but bounded route diversification. After that, it returned to a more conservative mode to consolidate the improvement. The final solution kept all deliveries assigned and respected capacity, time and priority constraints.
\end{quote}

This explanation avoids low-level algorithmic jargon while preserving the reasoning chain: what the agent observed, what memory or pattern it used, what action it selected, why the action was bounded, how risk was controlled and whether the result preserved business constraints.

Explainability is not an add-on. It is part of the value of putting a reasoning agent above a metaheuristic.

\section{Discussion}

The main finding is not that a specific routing rule is universally good. The main finding is that RACL provides a viable method for producing useful control rules.

The experiment demonstrates that operational memory can support metaheuristic control; reasoning can go beyond direct memory-derived rules; bounded interventions can be tested without changing business constraints; risk can be converted into guardrails; useful behavior can be consolidated into a reproducible policy; and the agent can explain its decisions in business-readable terms.

\begin{table}[t]
\centering
\caption{Scope of the current evidence.}
\label{tab:scope}
\small
\begin{tabular}{p{0.45\linewidth}p{0.45\linewidth}}
\toprule
Supported by the experiment & Outside the current claim \\
\midrule
A reasoning agent can use operational memory to derive useful control behavior over a metaheuristic. & The experiment does not claim that the specific discovered actions are universal routing rules. \\
The agent can formulate additional hypotheses, test bounded interventions and consolidate a stronger policy when evidence is favorable. & The experiment does not claim that RACL dominates every adaptive or hyper-heuristic baseline. \\
The resulting control policy can improve the fixed optimizer in the studied repeated local-routing setting while preserving feasibility. & The experiment does not claim that a fully unattended production deployment has already been implemented. \\
RACL decisions can be logged and translated into business-readable explanations. & The experiment does not claim that business users should tune algorithmic internals directly. \\
\bottomrule
\end{tabular}
\end{table}

This changes the role of the optimizer. The metaheuristic remains the engine that constructs feasible solutions, but RACL becomes a learning layer around it. The optimizer no longer has to remain fixed after initial configuration; its control behavior can evolve as operational memory grows.

The practical value is especially clear for organizations that repeatedly solve similar optimization problems but lack internal optimization specialists. RACL offers a path toward continuous improvement without asking the business user to understand or tune metaheuristic internals.

\section{Scope and Future Work}

The current experiment validates the RACL cycle through a Codex-in-the-loop reasoning process and a reproducible policy proxy. A full unattended production deployment would call a reasoning model during execution blocks and update memory continuously.

The current routing testbed is intentionally concrete, but RACL is not tied to this specific ALNS-style implementation. Future work should test the same method with other metaheuristics, broader benchmarks, stronger adaptive baselines and live LLM/API execution.

The next research steps are formalizing train/validation/test memory splits, adding ablations for memory, guardrails and consolidation, comparing against stronger adaptive operator-selection baselines, validating business-facing explanations with operational users and testing whether rules can be retired or narrowed when later evidence contradicts them.

\section{Conclusion}

This paper proposes RACL, a Reasoning-Agent Control Layer for continuous metaheuristic learning.

RACL places a reasoning agent above an existing optimizer. The agent observes operational memory, reasons about search behavior, proposes bounded interventions, evaluates outcomes, applies guardrails, consolidates useful policies and explains its decisions. It improves the optimizer's algorithmic control behavior without modifying business constraints.

The routing experiment shows that this process can produce useful control behavior: RACL improves the operational memory policy in 18 of 21 feasible cases, improves or ties a non-reasoning stagnation-triggered policy in 18 of 21 feasible cases, and improves the fixed baseline in the measured paired sample.

The contribution is the method, not the specific rule. RACL demonstrates a path toward metaheuristics that learn from their own operational history through a reasoning agent placed at the control layer.

\bibliographystyle{plainnat}
\bibliography{references}

\end{document}